# On the Number of Samples Needed to Learn the Correct Structure of a Bayesian Network

**Or Zuk**
Dept. Phys. Comp. Systems
Weizmann Inst. of Science
Rehovot, 76100, Israel
*or.zuk@weizmann.ac.il*

**Shiri Margel**
Dept. Phys. Comp. Systems
Weizmann Inst. of Science
Rehovot, 76100, Israel
*shiri.margel@weizmann.ac.il*

**Eytan Domany**
Dept. Phys. Comp. Systems
Weizmann Inst. of Science
Rehovot, 76100, Israel
*eytan.domany@weizmann.ac.il*

## Abstract

Bayesian Networks (BNs) are useful tools giving a natural and compact representation of joint probability distributions. In many applications one needs to learn a Bayesian Network (BN) from data. In this context, it is important to understand the number of samples needed in order to guarantee a successful learning. Previous works have studied BNs sample complexity, yet they mainly focused on the requirement that the learned *distribution* will be close to the original distribution which generated the data. In this work, we study a different aspect of the learning task, namely the number of samples needed in order to learn the correct *structure* of the network. We give both asymptotic results (lower and upper-bounds) on the probability of learning a wrong structure, valid in the large sample limit, and experimental results, demonstrating the learning behavior for feasible sample sizes.

## 1 Introduction

Bayesian Networks (BNs) are generative models which are well suited for representing knowledge under uncertainty. Their compact representation, modularity and intuitive causal interpretation has made them very popular, and today they are used in various fields such as AI, Expert systems, Economics and Computational Biology. A BN is specified by two components: a qualitative part (structure), which represents conditional (in)dependencies between the random variables (r.v.s.), and a quantitative part (parametrization), representing the exact joint probability distribution of related r.v.s. in the network. Often there is no way of building BNs according to expert knowledge. In these cases one needs to learn a BN from data, i.e. from a sample of $N$ realizations (values taken by the network's variables), and it is important to assess the sample size one needs in order to learn a BN that approximates the true one with given quality. The subject of sample complexity of BNs has drawn some attention in the past decade. The *PAC* framework (Valiant 1984) was used in (Dasgupta 1997) for BNs with known structure, with and without hidden variables, and by various authors (Abbeel et al. 2005; Friedman and Yakhini 1996; Hoffgen 1993), for discrete BNs with unknown structure. All of these works used the *Kullback-Leibler* distance (Kullback and Leibler 1951) (also known as relative entropy) between the original and learned model's *distributions* to measure the approximation quality. A different criterion was presented in (Greiner et al. 1997), who measured the performance of the learned model (compared to the original) when answering queries (which depends, of course, on the query distribution). Our study is concentrated on a different aspect of the learning task, which is the number of samples needed in order to learn the correct network *structure*.

Intuition and practice indicate that learning the correct structure[1] is a much harder task than approximating the original distribution, and hence requires a larger number of samples. In (Dai et al. 1997) the relation between the number of causal errors (e.g. misidentified edges and edges directions) and sample size (as well as links strength) was studied using computer simulations for two specific learning algorithms. We derive rigorous upper and lower-bounds on the error probability, as well as experimental results, which do not depend on the learning algorithm used.

---

[1] When we refer to the correct structure, we actually mean the correct *equivalence class*, as the specific structure within the equivalence class cannot be distinguished based on observational data solely.

For simplicity, we concentrate on BNs with boolean r.v.s, yet our techniques can be applied to any discrete-node BNs. We give bounds on the error probability, i.e. the probability of learning a wrong structure, for the Bayesian-Information-Criteria (BIC), which is a specific case of the Minimum-Description-Length (MDL) score. Our results can be easily modified to give the asymptotic behavior for MDL scores with other penalty terms. We obtain different bounds for structures which cannot represent the true distribution ('under-fitting'), and structures which do represent the true distribution yet do not have the minimal number of parameters ('over-fitting'). For these two tasks we use, respectively, large and moderate deviations results. While these results are correct in the large sample limit, we also give experimental results for a moderate number of samples. We address the problem of learning the exact *correct* BN structure (up to equivalence classes considerations), and do not consider the case of learning a structure that approximates the original one. A relaxation of this requirement (for example allowing a certain fraction of mis-learned edges), leads to lower sample complexity. On the other hand, we assume that the learner is computationally unbounded and learns by scoring all possible models exhaustively and selecting the best scoring one. Obviously, the number of samples needed by real-life algorithms for structure learning may be higher.

In the next section we give basic definitions and results for BNs, MDL score, and relative entropy, which are used later. In section 3 we give asymptotic sample complexity bounds for learning the correct structure. In section 4 we present computer simulations demonstrating the behavior of the error probability for a moderate number of samples. We end with conclusions and future directions.

## 2 Preliminaries and Definitions

Let $X_1,..,X_n$ be binary random variables (r.v.s), with joint probability distribution $P$. In general, uppercase will denote random variables, and lowercase will denote their realizations. We will sometimes omit the latter, so for example $P(x_1,..,x_n) = P(X_1 = x_1,..,X_n = x_n)$. Formally, a Bayesian Network (BN) is a pair $B = \langle G, \Theta \rangle$. Here $G = \langle V, E \rangle$ is an $n$-vertex *directed acyclic graph (DAG)*, called the *structure*, which represents (in)dependencies between the $X_i$'s, such that a variable is independent of its non-descenders, given its parents (Pearl 1988). $\Theta = \{\Theta_i\}_{i=1}^n$ is the *parametriza-*

*tion*, specifying conditional probabilities, such that $\Theta_{i|pa_G(i)} = P_B(X_i = 1|Pa_G(i) = pa_G(i))$, where $Pa_G(i)$ are the parents of $X_i$ in the graph $G$. The set of parents of $X_j$ excluding $X_i$ is denoted $Pa_G(j \setminus i)$, and its values are denoted by $pa_G(j \setminus i)$. Keeping the other parent's values and setting $X_i = k$ will be denoted by $Pa_G(j \setminus i : k)$, for $k = 0, 1$.

The set of probability distributions on $m$ values is the $m-1$-dimensional simplex denoted $S_m$. The joint probability distribution associated with $B$ is denoted $P_B$, $P_B \in S_{2^n}$, and satisfies:

$$P_B(X_1,..,X_n) = \prod_{i=1}^n P_B(X_i|X_1,..,X_{i-1}) =$$

$$\prod_{i=1}^n P_B(X_i|Pa_G(i)) = \prod_{i=1}^n \Theta_{i|pa_G(i)}^{x_i}(1-\Theta_{i|pa_G(i)})^{1-x_i} \quad (1)$$

The graph dimension $|G| = \sum_{i=1}^n 2^{|Pa_G(i)|}$ is the number of parameters needed to specify $P_B$ when the structure is $G$. We assume that the correct BN belongs to a set of distributions which are "bounded-away from zero" with a bound $\gamma > 0$. This simply means

$$\gamma = \gamma(B) = \min_{i, pa_G(i)} \{ \min(\Theta_{i|pa_G(i)}, 1-\Theta_{i|pa_G(i)}) \}. \quad (2)$$

For any edge $(i,j) \in E$, its 'information content' in $B$ is defined as:

$$IC_B(i,j) = \min_{S \subset \{X_1,..,X_n\} \setminus \{X_i, X_j\}} I_{P_B}(X_i, X_j|S) \quad (3)$$

We also define $IC_B \equiv \min_{(i,j) \in E} IC_B(i,j)$. The parameters $\gamma$ and $IC_B$ (which depend on $B$) influence the sample complexity bounds we will obtain, and when they tend to zero, these bounds become large.

For a distribution $P$ we denote by $I(P)$ the set of all conditional independence relations of the form $X \perp Y|Z$ that hold in $P$, for $X, Y, Z$ disjoint sets of r.v.s. For a graph $G$ we denote by $I(G)$ all the independence assertions implied by $G$. $G$ is called an I-map for $P$ if $I(P) \supset I(G)$. $G$ is called a *P-map* for $P$ if $I(G) = I(P)$. We denote by $\mathcal{M}(G)$ the set of all distributions for which $G$ is an I-map, $\mathcal{M}(G) \subset S_{2^n}$. We will assume that our samples are generated from a probability distribution for which $G^*$ is a *P-map*. The BN generating the samples is denoted $B^* = \langle G^*, \Theta^* \rangle$, and the corresponding distribution is denoted $P_{B^*}$. $G^*$ is the unique *P-map* for $P_{B^*}$, up to graph-equivalent structures. Moreover, $|G^*| < |G|$ for any other *I-map* $G$ of $P$. We refer to $G^*$ as the 'correct' structure for $P_{B^*}$, and our purpose is to recover it (more precisely its equivalence class) from the data.

A fully-connected $DAG$ which is consistent with the ordering of $G^*$ is denoted $C^*$. The set of graphs obtained from fully-connected $DAGs$ by removing the edge $(i,j)$ (or $(j,i)$) is denoted $C_{ij}$, where the same notation is used for all such $DAGs$. We denote by $N$ the number of realizations (also referred to as samples) in the learning set. The samples are denoted by $x^{(i)}, i = 1,..,N$, with $x^{(i)} = (x_1^{(i)}, .., x_n^{(i)})$, so $x_j^{(i)}$ is the value of the r.v. $X_j$ in the $i$-th sample. We assume the samples are i.i.d., with $x^{(i)} \sim P_{B^*}$. For a distribution $P$ we denote by $P^{(N)}$ the joint product measure of $N$ i.i.d. r.v.s, each distributed according to $P$. Thus, for example, $(X^{(1)}, .., X^{(N)}) \sim P_{B^*}^{(N)}$. One can use a sample of size $N$ to estimate $P_{B^*}$ by simply counting the number of occurrences of each value $x \in \{0,1\}^n$ in the sample. The resulting distribution is called the sample distribution $\hat{P}_N$, given by $\hat{P}_N(x) = \frac{1}{N} \sum_{i=1}^N 1_{\{x^{(i)}=x\}}$.

For any $B = \langle G, \Theta \rangle$, the log-likelihood of the data is:

$$LL_N(G, \Theta) = \sum_{i=1}^N \sum_{j=1}^n \log P_B(x_j^{(i)} | pa_G^{(i)}(j)) = \quad (4)$$

$$\sum_{j=1}^n \sum_{i=1}^N \left[ x_j^{(i)} \log \Theta_{j|pa_G^{(i)}(j)} + (1-x_j^{(i)}) \log(1-\Theta_{j|pa_G^{(i)}(j)}) \right]$$

Given $G$, the maximal likelihood parametrization $\hat{\Theta} = \hat{\Theta}(G)$, is simply given by the sample probability, so $\hat{\Theta}_{i|pa_G(i)} = \hat{P}_N(X_i = 1 | pa_G(i))$. Using simple algebra, the likelihood of the model $G$ is given by:

$$LL_N(G) \equiv LL_N(G, \hat{\Theta}) = -N \sum_{j=1}^n H_{\hat{P}_N}(X_j | Pa_G(j)) \quad (5)$$

Where $H_P$ is the entropy or conditional entropy of a variable with respect to $P$. Similarly, $I_P$ denotes the mutual information (or conditional mutual information) of two variables. We denote $P_{N,G}$ the sample distribution $assuming$ that the correct structure is $G$, which is the probability distribution associated with the BN $\langle G, \hat{\Theta}(G) \rangle$. If $G$ is the best scoring structure, then $P_{N,G} = \hat{P}_N$.

A score is called asymptotically consistent if, when $N \to \infty$, the 'correct' model will attain the highest score with probability approaching one. Here, the 'correct' model refers to the graph (or actually its equivalence class) with minimal number of parameters which is an $I$-map for $P_{B^*}$. It can be easily shown that the log-likelihood is not a consistent score, and it is thus not useful for comparing structures. Adding edges to the graph always improves the likelihood, which makes the complete graph the highest scoring one, regardless of the true BN generating the data. A common strategy to cope with this overfitting problem, is based on the $Minimal\ Description\ Length\ (MDL)$ principle (Rissanen 1978). The $MDL$ score 'penalizes' complex models, thus giving a trade-off between data-fitting and model complexity. It usually takes the form:

$$S_N(G) = LL_N(G) - |G|\Psi(N) \quad (6)$$

where $\Psi \equiv \Psi(N)$ is a $penalty\ function$. We assume that $\Psi$ satisfies:

$$\lim_{N \to \infty} \Psi(N)^{-1} = \lim_{N \to \infty} \frac{\Psi(N)}{N} = 0. \quad (7)$$

Under these assumptions, the $MDL$ score is asymptotically consistent (Haughton 1988, 1989). Of particular interest is the choice $\Psi(N) = \frac{1}{2} \log N$, since in this case the score in eq. 6 is known to be asymptotically equivalent to the Bayesian score (with any nowhere-vanishing prior), and is also termed the $Bayesian\ Information\ Criterion\ (BIC)$ (Schwarz 1978).

### 2.1 Relative Entropy Properties

Working with the relative entropy distance measure often possesses technical difficulties, since it does not satisfy the requirements of a norm. Here we prove two results for the relative entropy, analogous to the symmetry and triangle inequality properties of a norm. Our results rely on the fact that the reference distribution is bounded away from zero, and involve constants which depend on this proximity to zero.

Let $P, Q$ be two strictly positive discrete probability distributions over $m$ values. Their relative entropy is:

$$D(P||Q) = \sum_{i=1}^m P_i \log \frac{P_i}{Q_i} \quad (8)$$

For a graph $G$ and a probability $P$ we denote $D(G||P) = \inf_{Q \in \mathcal{M}(G)} D(Q||P)$. We can always write $Q$ uniquely as $Q = P + \epsilon V$, where $\epsilon > 0$ and $V$ is a unit vector. Moreover, since $P$ and $Q$ are probability distributions, $V$ also satisfies: $\sum_{i=1}^N V_i = 0$. Taking Taylor expansion of the relative entropy gives:

$$D(P||P+\epsilon V), D(P+\epsilon V||P) = \frac{1}{2}\epsilon^2 \sum_{i=1}^m \frac{V_i^2}{P_i} + O(\epsilon^3) \quad (9)$$

We see that the relative entropy behaves locally as $O(\epsilon^2)$. Moreover, it is 'locally symmetric', in the sense that $\frac{D(P||P+\epsilon V)}{D(P+\epsilon V||P)} \to 1$ as $\epsilon \to 0$. We can also formulate the following $uniform$ bounds, independent of $V$'s direction:

**Lemma 1** *Let $P, Q$ be two distributions on $m$ values, with $\gamma = \min_i P_i > 0$ and $||P - Q||_2 = \epsilon < \frac{\gamma}{2}$. Then $D(P||Q), D(Q||P) \leq \frac{8\epsilon^2}{\gamma^2}$.*

**Proof** Write $Q = P + \epsilon V$ where $V$ is a unit vector. Using Taylor expansion we get:

$$D(P||Q) = D(P||P + \epsilon V) = \sum_{k=2}^{\infty} \frac{(-\epsilon)^k}{k} \sum_{i=1}^{m} \frac{V_i^k}{P_i^{k-1}} \leq$$

$$\sum_{k=2}^{\infty} \epsilon^k \sum_{i=1}^{m} \left|\frac{V_i}{P_i}\right|^k \leq \sum_{k=2}^{\infty} \left(\frac{\epsilon}{\gamma}\right)^k = \frac{\epsilon^2}{\gamma(\gamma - \epsilon)} \leq \frac{2\epsilon^2}{\gamma^2} \quad (10)$$

Where convergence is guaranteed since $\frac{\epsilon}{\gamma} < \frac{1}{2}$. For $D(Q||P)$, just flip between $P$ and $Q$, and since $\min_i Q_i \geq \frac{\gamma}{2}$, the bound is multiplied by four. ∎

**Lemma 2** *Let $P$ be a distribution on $m$ values, with $\gamma = \min_i P_i > 0$. Let $Q_1$ and $Q_2$ be any two distributions satisfying $D(Q_1||Q_2), D(Q_1||P) \leq \epsilon$ for some positive constant $\epsilon < \frac{\gamma^2}{32 \log 2}$. Then we also have $D(P||Q_1), D(Q_2||P), D(P||Q_2) \leq \frac{64 \log 2}{\gamma^2} \epsilon$.*

**Proof** We use the following inequality from (Cover and Thomas 1991):

$$D(P||Q) \geq \frac{1}{2 \log 2} ||P - Q||_1^2 \quad (11)$$

To get: $||Q_1 - Q_2||_1^2, ||Q_1 - P||_1^2 \leq 2\epsilon \log 2$. Since $\sqrt{2\epsilon \log 2} < \frac{\gamma}{2}$, we get from lemma 1 that $D(P||Q_1) \leq \frac{64 m \log 2}{\gamma^2} \epsilon$. From the triangle inequality (in $L_1$) we get:

$$||Q_2 - P||_1^2 \leq 2(||Q_1 - Q_2||_1^2 + ||Q_1 - P||_1^2) \leq 8\epsilon \log 2$$

But $\sqrt{8\epsilon \log 2} < \frac{\gamma}{2}$, therefore, using lemma 1 again, we get that $D(Q_2||P), D(P||Q_2) < \frac{64 \log 2}{\gamma^2} \epsilon$. ∎

## 3 Learning the Correct Structure

This section is devoted to the problem of identifying the correct network structure from data. This problem is also known in the statistics literature as 'model selection'. For exponential families, the problem was studied in (Haughton 1988), where asymptotic consistency was established, and in (Haughton 1989), where asymptotic results on the error-probability as a function of the number of samples were obtained. In (Geiger et al. 2001) it was shown that directed graphical models, with no hidden variables, are in fact curved exponential families, and thus Haughton's results are applicable for them. Nevertheless, the explicit constants appearing in Haughton's asymptotical analysis were not fully characterized. We will present these results in the context of learning BNs structure, and, when possible, give bounds on the constants governing the decay of the error probability.

The different graph structures, which are curved exponential families, can be thought of as Riemannian manifolds $\mathcal{M}(G) \subset S_{2^n}$ (Geiger et al. 2001) of different dimensions which are subsets of the simplex $S_{2^n}$, 'competing' for their fit to the data, which can be casted simply as their (Kullback-Leibler) distance from the sample probability $\hat{P}_N \in S_{2^n}$, where each such manifold 'pays' a penalty proportional to its dimension. While analyzing these models together seems complicated, our approach is studying the error probability of one model at a time, i.e. $P_{B^*}^{(N)}(S_N(G^*) < S_N(G))$. Following (Haughton 1988, 1989), we divide the models into two disjoint subsets: 1. Graphs $G$ which are not I-maps for $P_{B^*}$ (i.e. $P_{B^*} \notin \mathcal{M}(G)$), and 2. graphs $G$ which are I-maps for $P_{B^*}$, yet have higher dimension than $G^*$, $P_{B^*} \in \mathcal{M}(G)$ and $|G| > |G^*|$.

### 3.1 Graphs which are not I-maps for $P_{B^*}$

We begin by bounding the error in learning a graph which is not an *I-map* for $P_{B^*}$:

**Theorem 1** *Let $P_{B^*} \in \mathcal{M}(G^*)$ where $G^*$ is a P-map for $P_{B^*}$. Let $G$ be another graph which is not an I-map for $P_{B^*}$. Then $\exists c > 0$ such that:*

$$\limsup_{N \to \infty} \frac{1}{N} \log P_{B^*}^{(N)}(S_N(G^*) < S_N(G)) \leq -c. \quad (12)$$

*If, in addition, $|G| \leq |G^*|$, then:*

$$\liminf_{N \to \infty} \frac{1}{N} \log P_{B^*}^{(N)}(S_N(G^*) < S_N(G)) \geq$$
$$-D(G||P_{B^*}) \log 2 \quad (13)$$

**Proof** The first inequality follows directly from (Haughton 1989), proposition 2. For the second inequality, note that if $\hat{P}_N \in \mathcal{M}(G)$, then $LL_N(G^*) \leq LL_N(G)$, and therefore $S_N(G^*) < S_N(G)$. This event happens, according to the Sanov theorem[2] (Sanov 1957), with probability $\Omega(2^{-ND(G||P_{B^*})})$, which completes the proof. ∎

Note that for a given $P$ and $G$, it might be difficult to compute the relative entropy $D(G||P)$. However,

---

[2] Actually, one cannot use here directly the Sanov lower bound, since $\mathcal{M}(G)$ is typically a manifold of a lower dimension than $S_{2^n}$. One can overcome this technical difficulty by using an $\epsilon$-neighborhood of $\mathcal{M}(G)$, yet the proof's details are omitted due to lack of space.

the relative entropy in the other direction $D(P||G)$ is easily given by:

$$D(P||G) = \sum_{i=1}^{n} I_P\big(X_i, \{X_1, .., X_{i-1}\} \setminus Pa_G(i) | Pa_G(i)\big). \quad (14)$$

One can apply lemma 2, to get an explicit lower-bound on the error exponent in the above theorem. Unfortunately, theorem 1 does not give us an explicit upper-bound on the error exponent. The rest of this section is devoted to finding such a bound. Following (Friedman and Yakhini 1996), we derive our upper-bound in two steps. First, we assume that the observed distribution is *ideal*, that is $\hat{P}_N = P_{B^*}$. Although this assumption is unrealistic, and many times even not feasible, it helps us to understand the trade-off between data-fitting and model complexity in the *MDL* score. Next, we study the effect of sampling noise on our bound. By using concentration of measure arguments (specifically Chernoff bounds), we show that with high probability $\hat{P}_N$ is close to $P_{B^*}$, allowing us to get an upper-bound in the presence of sampling noise. Before deriving our results, we bring a useful relation for *P-map*s. From definition, if $G = \langle V, E \rangle$ is a *P-map* for $P$, then $I_P\big(X_i, X_j | Pa_G(j \setminus i)\big) > 0$ for any $(i,j) \in E$. For strictly positive BNs, the following stronger relation holds:

**Proposition 1** *Let $B = \langle G, \Theta \rangle$ be a BN with $G = \langle V, E \rangle$ a P-map for $P_B$ and $\Theta \in (0,1)$. Then $IC_B > 0$.*

The positivity of $IC_B$ means that no edge in $G$ is 'redundant' with respect to the distribution $P_B$. The magnitude of $IC_B$ can be thought of as a measure of how much information on $P_B$ is captured in $G$, compared to lower-dimensional models. Intuitively, one expect that the higher this value is, the easier it will be to separate $G$ from lower-dimensional models based on data sampled from $P_B$. The bounds obtained in the next section show indeed such dependence on $IC_B$, as well as a dependence on $\gamma$. Our purpose is not to achieve the tightest bounds possible, but merely to demonstrate the dependence of the convergence rate on these two parameters.

### 3.1.1 Deriving Bounds in the Ideal Case

In the ideal case all our statistics are deterministic, and depend only on the sample size. We denote them with the superscript $(I)$, thus, for example, the log-likelihood is denoted $LL_N^{(I)}$.

**Lemma 3** *In the ideal case, if $S_N^{(I)}(G) > S_N^{(I)}(G^*)$, then $|G| < |G^*|$.*

**Proof** Since $\hat{P}_N = P_{B^*}$, we get $LL_N^{(I)}(G^*) = LL_N^{(I)}(C^*)$. But $LL_N^{(I)}(G) \leq LL_N^{(I)}(C^*), \forall G$, therefore:

$$LL_N^{(I)}(G^*) - |G^*|\Psi(N) < LL_N^{(I)}(G) - |G|\Psi(N) \leq$$
$$LL_N^{(I)}(G^*) - |G|\Psi(N) \Rightarrow |G| < |G^*|. \quad (15)$$

∎

**Lemma 4**

$$S_N^{(I)}(G^*) \geq \max_{(i,j) \in E^*} LL_N^{(I)}(C_{ij}) - n\Psi(N) \Rightarrow$$
$$S_N^{(I)}(G^*) = \max_G S_N^{(I)}(G) \quad (16)$$

**Proof** Assume, negatively, that $\exists G$, $S_N^{(I)}(G^*) < S_N^{(I)}(G)$. From lemma 3 we get $|G| < |G^*|$, therefore we have some edge $(i,j) \in E^* \setminus E$. Thus, $\mathcal{M}(G) \subset \mathcal{M}(C_{ij})$ and $LL_N^{(I)}(G) \leq LL_N^{(I)}(C_{ij})$. Using the lemma's condition and $|G| \geq n$, we get $S_N^{(I)}(G^*) \geq LL_N^{(I)}(G) - n\Psi(N) \geq S_N^{(I)}(G)$, yielding a contradiction. ∎

The likelihood in the *ideal* case is given by:

$$LL_N^{(I)}(C^*) = LL_N^{(I)}(G^*) = -N \sum_{i=1}^{n} H_{P_{B^*}}\big(X_i | Pa_{G^*}(i)\big) \quad (17)$$

The likelihood loss, when removing an edge $(i,j)$, is:

$$LL_N^{(I)}(C^*) - LL_N^{(I)}(C_{ij}) =$$
$$N \cdot I_{P_{B^*}}\big(X_i, X_j | Pa_{C_{ij}}(i)\big) \geq N \cdot IC_{B^*}(i,j) \quad (18)$$

Which is positive by proposition 1. By using lemma 4 and eq. 18, we get:

**Proposition 2** *In the ideal case:*

$$\frac{\Psi(N)}{N} \leq \frac{IC_{B^*}}{|G^*| - n} \Rightarrow S_N^{(I)}(G^*) = \max_G S_N^{(I)}(G). \quad (19)$$

### 3.1.2 Treatment of the Noisy Case

Note that a weaker form of lemma 4 is still valid in the noisy case, assuming that we require only $S_N(G^*) \geq \max_{G, |G| \leq |G^*|} S_N(G)$. Our method of proof is showing that the likelihood difference $LL_N(G^*) - LL_N(C_{ij})$ in the noisy case, is close, with high probability, to the ideal version shown in eq. 18. In order to show this proximity, we use a series of concentration lemmas:

**Lemma 5** Let $B = \langle G, \Theta \rangle$ be a BN. Take a subset $S \subset \{X_1,..,X_n\}$ of r.v.s. Then $\forall \alpha \in (0, \gamma^n)$:

$$P_B^{(N)}\big(|\hat{P}_N(S) - P_B(S)| \geq \alpha\big) \leq 2e^{-\alpha^2 N/3P_B(S)} \quad (20)$$

**Proof** Let $Y_N = \sum_{j=1}^N 1_{\{S^{(j)}=s\}}$ be the r.v. counting the number of samples in which the value of the $S$ variables was $s$. Then $Y_N \sim Binomial(N, P_B(S))$. Using Chernoff bounds we get for $\alpha \in (0, 1)$,

$$P_B^{(N)}\big((P_B(S) - \alpha)N \leq Y_N \leq (P_B(S) + \alpha)N\big) \geq$$
$$1 - 2e^{-\alpha^2 N/3P_B(S)} \quad (21)$$

Noting that $Y_N = N\hat{P}_N(S)$ completes the proof. ∎

**Lemma 6** Let $B = \langle G, \Theta \rangle$ be a BN. Take a subset $S \subset \{X_1,..,X_n\}$ of r.v.s. Then $\forall \alpha \in (0, 1)$:

$$P_B^{(N)}\big(|\hat{P}_N(S) \log \hat{P}_N(S) - P_B(S) \log P_B(S)| \geq$$
$$\alpha[\log(\gamma^n - \alpha) + 1]\big) \leq 2e^{-\alpha^2 N/3P_B(S)} \quad (22)$$

**Proof** The function $f(x) = x \log x$ is Lipschitz continuous in the interval $[a, 1]$ with Lipschitz constant $\log a + 1$. From lemma 5, the inequality $|\hat{P}_N(S) - P_B(S)|$ holds with the desired probability. In that case, obviously we have $\gamma^n - \alpha \leq \hat{P}_N(S), P_B(S)$ and from here eq. 22 follows immediately. ∎

We now give concentration bounds on the empirical entropy and conditional entropy functions:

**Lemma 7** Let $B = \langle G, \Theta \rangle$ be a BN. Take two disjoint subsets $S, T \subset \{X_1,..,X_n\}$ of r.v.s. Then $\forall \alpha \in (0, \gamma^n)$:

$$P_B^{(N)}\big(|H_{\hat{P}_N}(S) - H_{P_B}(S)| \geq$$
$$2^n \alpha[\log(\gamma^n - \alpha) + 1]\big) \leq 2^{n+1} e^{-\alpha^2 N/3} \quad (23)$$

And:

$$P_B^{(N)}\big(|H_{\hat{P}_N}(T|S) - H_{P_B}(T|S)| \geq$$
$$2^{n+1} \alpha[\log(\gamma^n - \alpha) + 1]\big) \leq 2^{n+2} e^{-\alpha^2 N/3} \quad (24)$$

**Proof** For the first part, simply sum over all possible realizations of $S$ in lemma 6, and apply the union bound, noting that $P_B(S) < 1$. For the second part, use the chain rule for entropy $H(Y|X) = H(X,Y) - H(X)$. Apply the first part on the sets $S$ and $S \cup T$, and apply the union bound again to get the desired result. ∎

The next lemma bounds the difference in the log-likelihood gap $LL_N(G^*) - LL_N(C_{ij})$ between the ideal and the noisy case:

**Lemma 8** Let $B = \langle G, \Theta \rangle$ be a BN. Take a subset $S \subset \{X_1,..,X_n\}$ of r.v.s. Then $\forall \alpha \in (0, 1)$:

$$P_B^{(N)}\Big(|[LL_N(G^*) - LL_N(C_{ij})] -$$
$$[LL_N^{(I)}(G^*) - LL_N^{(I)}(C_{ij})]| \geq$$
$$Nn2^{n+2}\alpha[\log(\gamma^n - \alpha) + 1]\Big) \leq n2^{n+3} e^{-\alpha^2 N/3} \quad (25)$$

**Proof** Recall, from eq. 5, that the log-likelihood can be written as $LL_N(G) = -N \sum_{j=1}^n H_{\hat{P}_N}(X_j|Pa_G(j))$. Apply lemma 7 with $T = X_j$ and $S = Pa_G(j)$, for $G = G^*$ and $\forall j = 1,..,n$. This can be used, along with the union bound, to bound $|LL_N(G^*) - LL_N^{(I)}(G^*)|$. Similarly, taking $G = C_{ij}$ in lemma 7 is used to bound $|LL_N(C_{ij}) - LL_N^{(I)}(C_{ij})|$. Combining the bounds by using the triangle inequality gives eq. 25. ∎

We are now ready to prove the main result of this section, giving our asymptotic upper-bound on the error exponent:

**Theorem 2** If $G$ is not an I-map of $P_{B^*}$, then

$$\limsup_{N \to \infty} \frac{1}{N} \log P_{B^*}^{(N)}\big(S_N(G^*) < S_N(G)\big) \leq$$
$$\max\Big(-\frac{\gamma^{2n}}{6}, -\frac{IC_{B^*}^2}{48[n\log(\gamma/2) + 1]^2 4^n}\Big) \quad (26)$$

**Proof** Set $\alpha = \min\Big(\frac{\gamma^n}{2}, \frac{IC_{B^*}}{2^{n+2}[n\log(\gamma/2)+1]}\Big)$. From lemma 8 and eq. 18, we get:

$$P_B^{(N)}\Big(LL_N(G^*) - LL_N(C_{ij}) \geq \frac{N \cdot IC_{B^*}}{2}\Big) \leq$$
$$P_B^{(N)}\Big(|[LL_N(G^*) - LL_N(C_{ij})] -$$
$$[LL_N^{(I)}(G^*) - LL_N^{(I)}(C_{ij})]| \geq \frac{N \cdot IC_{B^*}}{2}\Big) \leq$$
$$n2^{n+3} e^{-\alpha^2 N/3} \quad (27)$$

Assume that $N$ is large enough such that $\frac{\Psi(N)}{N} \leq \frac{IC_{B^*}}{2(|G^*|-n)}$. For such $N$'s, we get:

$$P_B^{(N)}\Big(LL_N(G^*) - LL_N(C_{ij}) \geq \Psi(N)\big(|G^*| - n\big)\Big) \leq$$
$$n2^{n+3} e^{-\alpha^2 N/3} \quad (28)$$

By using the weaker form of lemma 4, and taking the union bound over all $C_{ij}$'s, we get:

$$P_{B^*}^{(N)}\big(S_N(G^*) < S_N(G)\big) \leq \binom{n}{2} n2^{n+3} e^{-\alpha^2 N/3} \quad (29)$$

Taking the logarithms of both sides and substituting the $\alpha$ value we have chosen completes the proof. ∎

### 3.2 Graphs which are *I-maps* for $P_{B^*}$

In the previous section we have seen that if a graph $G$ is not an I-map for $P_{B^*}$, then $\mathcal{M}(G)$ is 'bounded away' from $P_{B^*}$ in the simplex $S_{2^n}$. This was used to relate the error probability to a large deviation event. When the graph $G$ is an *I-map* for $P_{B^*}$, we have $P_{B^*} \in \mathcal{M}(G)$. The event of choosing an over-parameterized *I-map* is a moderate deviation event, and the next theorem characterizes its asymptotic probability:

**Theorem 3** Let $P_{B^*} \in \mathcal{M}(G) \cap \mathcal{M}(G^*)$ and assume $|G| > |G^*|$. Then $\exists d > 0$ such that $P_{B^*}^{(N)}(S_N(G^*) < S_N(G)) = O(N^{-d})$. Moreover, if $I(G^*) \subset I(G)$ then:

$$Pr\big(S_N(G^*) < S_N(G)\big) \sim$$
$$\frac{1}{\Gamma\big(\frac{|G|-|G^*|}{2}\big)} (\log N)^{\frac{1}{2}(|G|-|G^*|)-1} N^{-\frac{1}{2}(|G|-|G^*|)} \quad (30)$$

**Proof** Recall that our models are in fact curved exponential families. Then the first statement is a direct consequence of proposition 3 in (Haughton 1989). The second is due to (Woodroofe 1978, 1982). ∎

Note that for nested models, the asymptotic error depends only on $G^*$ and not on the specific parametrization determining $P_{B^*}$. Computer simulations we have performed did not find a dependence on the parametrization also for non-nested models. This is substantially different from the under-fitting error shown in the previous section, where we have seen that the error exponents depends on the parametrization. The constant $d$ is proportional (also in Haughton's result) to the difference in the dimensions $|G|-|G^*|$: The more over-parameterized $G$ is, the faster the decay of the error probability.

## 4 Experimental Results

In the previous section we noted a qualitative difference between the over and under-fitting errors. While the under-fitting error decays exponentially fast with $N$, the over-fitting probability decays slower, as a power of $N$. As we show now, this difference is relevant mainly in the large $N$ limit, while for small $N$'s, as one might expect intuitively, the situation is the opposite and under-fitting is more likely. We examined the errors in the learning process for a small BN with four r.v.s. There are 543 different *DAGs* on four nodes, divided into 185 equivalence classes. We chose $G^*$ to be the graph with the edge set $E^* = \{(1,2),(1,3),(2,4),(3,4)\}$, and the parametrization to be $\Theta^* : \{\Theta_1^* = 0.1, \Theta_{2|0}^* = 0.1, \Theta_{2|1}^* = 0.3, \Theta_{3|0}^* = 0.1, \Theta_{3|1}^* = 0.3, \Theta_{4|00}^* = 0.1, \Theta_{4|01}^* = 0.3, \Theta_{4|10}^* = 0.8, \Theta_{4|11}^* = 0.2\}$. There are 4 equivalence classes (including $G^*$) which are I-maps for $P_{B^*}$, and 181 which under-fit it. We also took two specific structures, $G_1$ with $E_1 = \{(1,2),(1,3),(1,4),(2,4),(3,4)\}$, which is an over-parameterized I-map for $P_{B^*}$, and $G_2$ with $E_2 = \emptyset$, which is not an I-map for $P_{B^*}$. Since the errors become rare events for large $N$, we had to use importance-sampling methods in order to estimate their probability. Rather than generating samples directly from $P_{B^*}$, we used other distributions $Q$, for which these events are more likely (for example, for $G_2$ we took $Q$s for which the r.v.s. are almost independent), and applied the appropriate correction needed to estimate the probability given $P_{B^*}$. The error probabilities are presented in figure 1. For $N < 2000$, the error probability is dominated by the under-fitting structure, while for larger $N$s the over-fitting probability is higher. Qualitatively similar results were also obtained for other networks we have started with and other choices of competing wrong graphs.

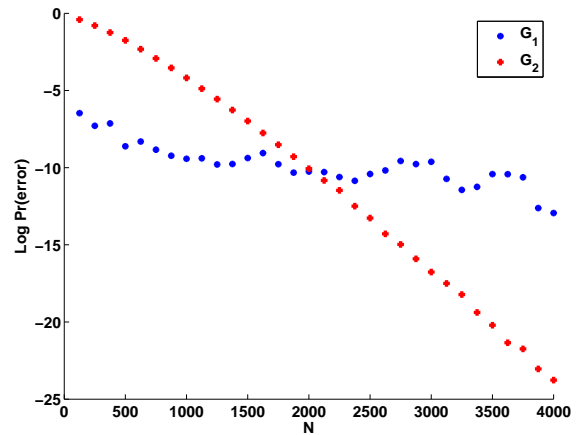

Figure 1: The log-error probabilities for learning two wrong graphs, shown as functions of the sample size for $G_1$ (in blue) which is an over-parameterized *I-map* for $P_{B^*}$, and for $G_2$ (in red) which is not an *I-map* for $P_{B^*}$. We used importance sampling and averaged over 30 different $Q$ distributions, where 6000 samples were drawn from each. The $G_2$ log-error exhibits an excellent fit for a straight line. The $G_1$ error is fitted better to a straight line on a loglog plot (not shown), in accordance with its power-law asymptotic behavior.

## 5 Discussion

We presented new lower and upper-bounds on the number of samples needed to ensure that the true and learned models share the same *structure*. When considering only graphs which are not I-maps for the correct distribution $P_{B^*}$, we have shown that the error probability decays exponentially with $N$. When allowing graphs with are I-maps for $P_{B^*}$ but are over-parameterized, we showed that the error probability decays as a power of $N$. While these results imply that in the large $N$ limit the error is dominated by over-fitting, for small values of $N$, the opposite situation of under-fitting the true model is more likely. This observation is corroborated by the fact that in few small BNs we have examined, the error exponent (both in our bounds and from numerical simulations) for the under-fitting case was close to zero, indicating a slow (although exponential) decay for moderate sample sizes. Although the BIC score has the desired property of consistency, perhaps a different choice of the penalty function (e.g. smaller than $\frac{1}{2}\log N$ for small values of $N$ and larger for large values of $N$), may lead to a smaller error probability.

The bounds we have obtained relate to the error of choosing a specific wrong model $G$, when its score is higher than that of the 'correct' model $G^*$. Typically, however, one needs to identify $G^*$ from a set of candidate models. Here one needs to look simultaneously on the (dependent) events $\{S_N(G^*) \leq S_N(G)\}$ for all possible wrong graphs $G$. A challenging future direction is trying to bound the error probability for learning any model from a set of wrong models together. We note that the number of possible candidate graphs grows super-exponentially with the network size $n$, even if we consider only bounded in-degree networks. Therefore, a simple union-bound argument may not give satisfactory results here, and one needs to turn into more sophisticated techniques, if one wants to study the dependence of sample complexity on the size of the network.

**Acknowledgements**

We thank E. Shimony and E. Segal for useful discussions and suggestions, and L. Hertzberg and L. Ein-Dor for reading the manuscript. This work was supported in part by grants from the Minerva Foundation, the Ridgefield Foundation and by the European Community's Human Potential Programme under contract HPRN-CT-2002-00319, STIPCO.

## References

P. Abbeel, D. Koller, and A.Y. Ng. Learning factor graphs in polynomial time and sample complexity. In *UAI*, pages 1–9, 2005.

T. M. Cover and J.M. Thomas. *Elements of Information Theory*. John Wiley, New York, NY, 1991.

H. Dai, K.B. Korb, C.S. Wallace, and X. Wu. A study of casual discovery with weak links and small samples. In *IJCAI*, pages 1304–1309, 1997.

S. Dasgupta. The sample complexity of learning fixed-structure bayesian nets. *Machine Learning*, 29(2-3): 165–180, 1997.

N. Friedman and Z. Yakhini. On the sample complexity of bayesian networks. In *UAI*, pages 274–282, 1996.

D. Geiger, D. Heckerman, H. King, and C. Meek. Stratified exponential families: Graphical models and model selection. *Ann. Statist.*, 29:505–529, 2001.

R. Greiner, A. Grove, and D. Schuurmans. Learning bayesian nets that perform well. In *UAI*, pages 198–207, Aug 1997.

D. M. A. Haughton. On the choice of a model to fit data from an exponential family. *Ann. Statist.*, 16: 342–355, 1988.

D. M. A. Haughton. Size of the error in the choice of a model to fit data from an exponential family. *Sankhya: Ind. J. Statist.*, 51:45–58, 1989.

K.U. Hoffgen. Learning and robust learning of product distributions. In *COLT*, pages 77–83, 1993.

S. Kullback and R. A. Leibler. On information and sufficiency. *Ann. Math. Statist.*, 22:79–86, 1951.

J. Pearl. *Probabilistic reasoning in Intelligent Systems: Networks of Plausible Inference*. Morgan Kaufman, San Francisco Calif., 1988.

J. Rissanen. Modeling by shortest data description. *Automatica*, 14:465–471, 1978.

I.N. Sanov. On the probability of large deviations of random variables. *Mat. Sbornik.*, 42:11–44, 1957.

G. Schwarz. Estimating the dimension of a model. *Ann. Statist.*, 6(2):461–464, 1978.

L. Valiant. A theory of the learnable. *Communications of the ACM*, 27:1134–1142, 1984.

M. Woodroofe. Large deviations of the likelihood ratio statistics with applications to sequential testing. *Ann. Statist.*, 6:72–84, 1978.

M. Woodroofe. On model selection and the arcsine laws. *Ann. Statist.*, 10:1182–1194, 1982.